*Commentary*

# The Case for Durative Actions: A Commentary on PDDL2.1

**David E. Smith**  DESMITH@ARC.NASA.GOV
*NASA Ames Research Center*
*Computational Sciences Division, Mail Stop: 269-2*
*Moffett Field, CA 94035, U.S.A.*

## Abstract

The addition of durative actions to PDDL2.1 sparked some controversy. Fox and Long argued that actions should be considered as instantaneous, but can start and stop processes. Ultimately, a limited notion of durative actions was incorporated into the language. I argue that this notion is still impoverished, and that the underlying philosophical position of regarding durative actions as being a shorthand for a start action, process, and stop action ignores the realities of modelling and execution for complex systems.

## 1. Introduction

PDDL2.1 introduces a limited notion of time into the classical STRIPS planning framework. In particular, it introduces the notion of *durative actions*, that is, actions that take time. However, the notion of durative action is rather limited, and somewhat begrudging. This reflects an underlying philosophical position by Fox and Long that actions are really instantaneous, but can initiate and terminate continuous processes. According to this view, durative actions are seen as a shorthand for a start action, process, and stop action. As a result, durative actions lack some important features, namely the ability to require that (pre)conditions hold over specified intervals, and that effects can take place at arbitrary time points within the action. Fox and Long have argued that these features can be captured by breaking up a durative action into a series of smaller actions that only have effects at the beginning and end, and only have preconditions at the beginning, end, and over the entire action. However, this representation is exceptionally cumbersome, and ignores the fact that an agent may not have separate control over these actions. In addition, this representation forces a planner to do additional work in order to connect the actions.

## 2. An Example

To illustrate the problems with the PDDL2.1 notion of durative action, consider a simple example of a spacecraft that must turn in order to point an instrument at a particular target. In order to turn the spacecraft, thrusters in the reaction control system (RCS) are fired in order to supply angular velocity. The spacecraft then coasts until it is pointing in the correct direction (or nearly so), when the RCS thrusters again fire in order to stop the rotation. Firing the thrusters consumes propellant, and requires that the controller be dedicated to the task. In addition, when the thrusters are firing, there is vibration of the spacecraft, so certain other operations cannot be performed. While the thruster firings are relatively quick, the coasting phase is not. In general, turning a large spacecraft is a slow process that may take several minutes. The reason is that speedy turns require greater acceleration and deceleration, and therefore consume more propellant.

The first question we need to answer is, what is the best way to model this complex operation? We could model the turning operation as an initial action to start the spacecraft turning, and another action to stop the turn, interspersed with "processes" that model what the craft is doing in between. At some level of detail, this seems to be a reasonable model of the physics. However, it may very well be that turning and guidance have been built in as primitive operations on the spacecraft, and there is no possibility of starting and stopping turns independently. We could then model the operation as consisting of an instantaneous action to start the turn, followed by a finite process that terminates when the turn is complete. But why bother? The fact is, we are interested in the effects of the process, which can only be initiated by starting the turn. For these reasons, it seems natural and proper to regard this as a "durative action", with effects that take place throughout the action.

Now let's suppose that we want to model this operation as a durative action in PDDL2.1. We could say something like:

```
(:durative-action turn
 :parameters  (?current-target ?new-target - target)
 :duration    (= ?duration (/  (angle ?current-target ?new-target)¹
                               (turn-rate)))
 :condition   (and  (at start (pointing ?current-target))
                    (at start (>= (propellant) propellant-required))
                    (at start (not (controller-in-use))))
 :effect      (and  (at start (not (pointing ?current-target)))
                    (at start (decrease (propellant) propellant-required))
                    (at start (controller-in-use))
                    (at start (vibration))
                    (at end (not (controller-in-use)))
                    (at end (not (vibration)))
                    (at end (pointing ?new-target))))
```

However, this model of the action is quite conservative. It ties up the controller for the entire turn operation, and specifies that vibration is present for the entire operation. In addition, it consumes all the required propellant at the beginning of the operation. In reality, the RCS is only firing at the beginning and end of the turn. As a result, the controller is only needed during those two periods, vibration is only present during those two periods, and the propellant is consumed during those two periods. This might not matter if the coast phase were relatively quick. However, as we indicated earlier, turning a large spacecraft can take several minutes. Unfortunately, PDDL2.1 has a rather limited notion of a durative action – we cannot specify action conditions or effects at times other than the start or end of the action.

## 3. Decomposition into Sub-actions

Fox and Long have pointed out[2] that it is possible to model a durative action with such intermediate conditions and effects by breaking it up into a sequence of sub-actions. For the turn action we would need three sub-actions as illustrated below: a start-turn action, a coast action, and a stop-turn action, together with a turn action to bind them all together.

---

1. Note that this assumes we have precomputed and provided the angles between all possible pairs of targets. If we do not want to do this, we must provide the (vector) direction for each target and the planner would need the ability to do vector arithmetic or trigonometry within formulas.
2. Personal communication.

```
(:durative-action turn
 :parameters  (?current-target ?new-target - target)
 :duration    (= ?duration (/  (angle ?current-target ?new-target)
                               (turn-rate)))
 :condition   (and  (at start (pointing ?current-target))
                    (at start (>= (propellant) propellant-required))
                    (at end (finished)))
 :effect      (and  (at start (not (pointing ?current-target)))
                    (at start (turning))
                    (at start (enabled-start-turn))
                    (at end (not (turning)))
                    (at end (not (finished-turning)))
                    (at end (pointing ?new-target))))

(:durative-action start-turn
 :parameters  ()
 :duration    (= ?duration (start-turn-duration))
 :condition   (and  (at start (not (controller-in-use)))
                    (at start (>= (propellant) (/ propellant-required 2)))
                    (over all (turning))
                    (over all (enabled-start-turn)))
 :effect      (and  (at start (decrease (propellant) (/ propellant-required 2)))
                    (at start (controller-in-use))
                    (at start (vibration))
                    (at end (not (controller-in-use)))
                    (at end (not (vibration)))
                    (at end (not (enabled-start-turn)))
                    (at end (enabled-coast))))

(:durative-action coast
 :parameters  ()
 :duration    (= ?duration (coast-duration))
 :condition   (and  (over all (turning))
                    (over all (enabled-coast)))
 :effect      (and  (at end (not (enabled-coast)))
                    (at end (enabled-stop-turn))))

(:durative-action stop-turn
 :parameters  ()
 :duration    (= ?duration (RCS-duration))
 :condition   (and  (at start (not (controller-in-use)))
                    (at start (>= (propellant) (/ propellant-required 2)))
                    (over all (turning))
                    (over all (enabled-stop-turn)))
 :effect      (and  (at start (decrease (propellant) (/ propellant-required 2)))
                    (at start (controller-in-use))
                    (at start (vibration))
                    (at end (not (controller-in-use)))
                    (at end (not (vibration)))
                    (at end (not (enabled-stop-turn)))
                    (at start (finished))))
```

Figure 1 shows graphically how these actions are tied together. If the goal is to be pointing at a particular target, a turn action will be required. The turn action has an end precondition of (finished), which can only be satisfied by adding a stop-turn action[3]. Stop-turn has an "over all" condition (enabled-stop-turn) that can only be satisfied by the end effect of a coast action. Likewise, the coast action has an "over all" condition (enabled-coast) that can only be satisfied by an end effect of a start-turn action. The start-turn action has an "over all" condition (enabled-start-turn) that can only be satisfied by a start effect of the turn action. As a result, the turn action forces all three sub-actions into the plan, and each sub-action forces its predecessor sub-actions and a turn action into the plan. All three of these sub-actions have an "over all" condition (turning) that is only satisfied during a turn action. As a result, the only way that all of this can be consistently achieved is if all three sub-actions are packed sequentially into the turn action.

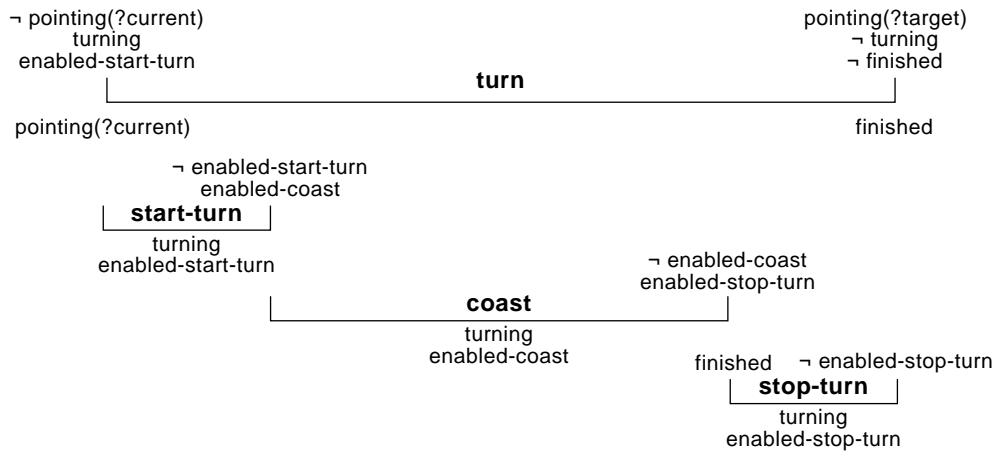

Figure 1: Sub-actions for the Turn operation. Start, end, and over-all conditions are shown below each action. The interconnecting start and end effects are shown above each action. For simplicity, I have omitted the effects concerning vibration, controller use, and propellant usage

There are two additional subtleties in this representation. The first is that, although each of the three sub-actions can only occur during a turn action, there is no obvious requirement that they occur during the *same* turn action. Suppose that we tried to place the start-turn action during a previous turn action. That previous turn action would have its own three sub-actions, and our wayward start-turn action would conflict with those sub-actions. Thus, in order to make this work, we would have to push those three sub-actions to an earlier turn action, and so on. Realizing that this cannot work requires a difficult induction argument. It seems unlikely that any existing planner could actually infer this, other than by trial and error. As a result, the process of generating plans involving such actions would incur a significant computational overhead, and engage in needless search.

A second subtlety that we have overlooked in this decomposition is that computing the durations of the sub-actions is a bit tricky. While it is reasonable to assume that the start and stop turn actions have fixed duration, the duration of the coast action depends on the current and target orientations of the spacecraft. In fact, the duration of the coast action must be the duration of the turn action minus the durations of the start and stop turn actions. The only way to do this is to introduce an additional numeric "turn-duration" function that is set by the turn action, and used to compute the duration of the coast action.

---

3. It turns out that the (finished) effect of stop-turn must be a start effect rather than an end effect. The reason is that if it were to occur as an end effect, the stop-turn action would need to complete prior to the end of the turn action, since the (finished) effect is mutex with the (not finished) end effect of the turn action. Despite this asymmetry in the representation, stop turn is still forced to occur wholly within the turn action because of the overall condition (turning).

So why is this process of decomposing an action into sub-actions so complex and convoluted? After all, in the HTN planning paradigm this is done all the time. The reason is that in generative planning we have adopted the view, for better or worse, that one is not allowed to directly specify how an action is to be used or how actions are connected with each other. As a result, in order to force the sub-actions to abut and fit within the turn action, we must do some tricky things. One might argue that we need this HTN capability in order to model such actions. Indeed, it would certainly make things easier. However, there is another way.

## 4. Richer Durative Actions

One approach to dealing with the above modelling problem is to admit a richer language for modelling durative actions. To make it convenient to model actions like the turn action, we need to be able to specify conditions that must hold at various points and intervals within the action, and effects that take place at various points and intervals within the action. There are many possible ways in which one could express such conditions and effects, but here is one straw-man possibility:

```
(:durative-action turn
 :parameters  (?current-target ?new-target - target)
 :duration    (= ?duration (/ (angle ?current-target ?new-target) (turn-rate)))
 :condition   (and  (at start (pointing ?current-target))
                    (at start (>= (propellant) propellant-required))
                    (at start (not (controller-in-use)))
                    (at (- end RCS-duration) (>= (propellant) (/ propellant-required 2)))
                    (at (- end RCS-duration) (not (controller-in-use))))
 :effect      (and  (at start (not (pointing ?current-target)))
                    (at start (decrease (propellant) (/ propellant-required 2)))
                    (over [start (+ start RCS-duration)] (controller-in-use))
                    (over [start (+ start RCS-duration)] (vibration))
                    (at (- end RCS-duration) (decrease (propellant) (/ propellant-required 2)))
                    (over [(- end RCS-duration) end] (controller-in-use))
                    (over [(- end RCS-duration) end] (vibration))
                    (at end (pointing ?new-target))))
```

Here we did not need to explicitly construct actions for starting and stopping the turn, or coasting. For this reason, we did not need to worry about their durations or about connecting these sub-actions. Instead, we simply specified the effects at the appropriate times during the turn action. Note that I specified vibration and controller use as interval effects. This seems less cumbersome than specifying two separate effects stating that the controller is in use at the beginning of an interval, and no longer in use at the end. However, there is also a more fundamental difference between the two encodings: in the encoding above, there is no possibility that another independent action could somehow make the controller available during the interval in which it is in use. Bedrax-Weiss et. al. (2003) have argued for the introduction of an explicit notion of resource into the PDDL language. If we had such a notion we could simplify the above encoding even further, by specifying that the controller is a reusable resource that is required by the turn action over the appropriate intervals. Vibration (or stability) could also be treated as a resource, although it is somewhat less intuitive to do so.

One final issue that we have avoided is the notion of continuous change. In our spacecraft example, there is certainly continuous change going on. Propellant is not burned instantaneously, and the orientation of the spacecraft changes continuously. The question is, do we need to model this? Certainly there are domains where it is necessary to reason about continuous change. As Fox and Long point out, when there are concurrent actions as well as simultaneous consumption and production of resources, it may be necessary to reason about how these resources change over the course of the actions. For example, a Mars rover receives energy from the solar panels at the same time it is driving from place to place. Since the battery has both a minimum and maximum capacity, one cannot model this easily using discrete consumption and production effects. However, if consumption and production do not happen simultaneously, one can model continuous change as taking place at the start or end of an action. This is sufficient for our spacecraft example since there are no actions that increase propellant, and one cannot perform two simultaneous actions that both affect the spacecraft's orientation.

## 5. Conclusion

Durative actions in PDDL2.1 are limited, and expressing complex durative actions by decomposition into sub-actions is difficult and clumsy. At the same time, it is not clear that modelling actions like turning a spacecraft in terms of processes is either necessary or useful, particularly when there is no possibility that the process can be deliberately interrupted. For domains like this, a richer, more expressive notion of durative action seems like the right modelling tool. Note that I would not argue that the modelling of processes is completely unnecessary. However, for many practical planning applications it is overkill. It results in a more complex representation and planning process than is necessary.

Is it cheating to model complex processes as durative actions? Of course it is. All modelling is cheating. In the real world of physics, nothing is instantaneous or indivisible, so it is cheating to model anything as an instantaneous action. Yet, we are usually content to model an action like turning on a light switch as instantaneous and indivisible, even though it does take a small amount of time, and there are complex processes behind it. A durative action is no different – we are simply choosing not to model the details of the process structure behind the action, even though it may be necessary to model the fact that the action takes time, and that the effects take place at different times during the action. For many practical applications, durative actions are an essential modelling tool, and they deserve a richer treatment than that provided in PDDL2.1.